\documentclass[sigconf,nonacm, natbib=false]{acmart}
\AtBeginDocument{%
  }

\RequirePackage[
  datamodel=acmdatamodel,
  style=acmnumeric,
  ]{biblatex}

\usepackage{xcolor}

\begin{document}


\title{Training Data Attribution for Image Generation using Ontology-Aligned Knowledge Graphs}
\renewcommand{\shorttitle}{Training Data Attribution for Image Generation using Ontology-Aligned Knowledge Graphs}


\author{Theodoros Aivalis}
\authornote{Corresponding Author. ORCID: 0009-0005-4452-9402}
\email{teoaivalis@iit.demokritos.gr}
\affiliation{%
  \institution{National Centre for Scientific Research ``Demokritos''}
  \country{Greece}
}
\affiliation{%
  \institution{University of Glasgow}
  \country{UK}
}

\author{Iraklis A. Klampanos}
\authornote{ORCID: 0000-0003-0478-4300}
\affiliation{%
  \institution{University of Glasgow}
  \country{UK}
}

\author{Antonis Troumpoukis}
\authornote{ORCID: 0000-0003-1078-8121}
\affiliation{%
  \institution{National Centre for Scientific Research ``Demokritos''}
  \country{Greece}
}

\author{Joemon M. Jose}
\authornote{ORCID: 0000-0001-9228-1759}
\affiliation{%
  \institution{University of Glasgow}
  \country{UK}
}

\begin{abstract}
As generative models become powerful, concerns around transparency, accountability, and copyright violations have intensified. Understanding how specific training data contributes to a model’s output is critical. We introduce a framework for interpreting generative outputs through the automatic construction of ontologyaligned knowledge graphs (KGs). While automatic KG construction from natural text has advanced, extracting structured and ontology-consistent representations from visual content remains challenging—due to the richness and multi-object nature of images. Leveraging multimodal large language models (LLMs), our method extracts structured triples from images, aligned with a domainspecific ontology. By comparing the KGs of generated and training images, we can trace potential influences, enabling copyright analysis, dataset transparency, and interpretable AI. We validate our method through experiments on locally trained models via unlearning, and on large-scale models through a style-specific experiment. Our framework supports the development of AI systems that foster human collaboration, creativity and stimulate curiosity.\footnote{A short version appears in the Proceedings of the 41st ACM/SIGAPP Symposium on Applied Computing (SAC’26).}
\end{abstract}

\keywords{Generative AI, Interpretability, KGs, Multimodal LLMs, Copyrights}

\maketitle
\section{Introduction}
\label{sec:introduction}
The rise of generative AI has intensified debate over its transparency and social impact. These models typically operate as opaque systems, offering little insight into how specific training data influences generated outputs. This lack of transparency poses serious challenges in domains where attribution, authorship, and legal rights matter, including artistic production, cultural heritage, and copyright-sensitive applications.

Legal and ethical concerns surrounding generative models are growing, particularly around the use of copyrighted content in model training. At the heart of these concerns lies the question of authorship and moral rights. The right to be recognised as the author of one’s work, described by Ginsburg as “the most moral of rights”  \cite{ginsburg2016moral}, is central to ongoing debates. A number of lawsuits argue that training on copyrighted content without consent violates property rights. For example, the New York Times has filed suit against OpenAI for unauthorised use of its journalistic archives\footnote{\url{https://harvardlawreview.org/blog/2024/04/nyt-v-openai-the-timess-about-face/}, as viewed December 2025.}, while Getty Images has taken legal action against Stability AI for generating content derived from its proprietary database\footnote{\url{https://www.theverge.com/2023/1/17/23558516/ai-art-copyright-stable-diffusion-getty-images-lawsuit}, as viewed December 2025.}. Beyond lawsuits, artists and performers have raised concerns. Indicatively, Elton John and Dua Lipa, along with 200 other artists, signed an open letter urging tech companies to protect human creativity from AI tools that could mimic their voices and styles without consent\footnote{\url{https://www.bbc.com/news/articles/c071elp1rv1o}, as viewed December 2025.}.

In response, policy bodies and international organisations proposed new governance frameworks. The European Union's AI Act  \cite{euaiact} outlines rules for transparency in AI systems, while the UK’s National AI Strategy  \cite{uk_aistrategy} and UNESCO’s ethical guidelines  \cite{unesco_ai} advocate for responsible AI, grounded in privacy and property protections. 
Recently, Denmark proposed an amendment to its copyright law that would grant individuals legal rights over their voice, face, and bodily likeness. Described as the first legislation of its kind in Europe, it seeks to address legal gaps around AI-generated imitations of personal characteristics. As the Danish culture minister noted, it ensures “the right to their own body, voice and facial features,” underscoring the need for identity protections over generative AI\footnote{\url{https://www.theguardian.com/technology/2025/jun/27/deepfakes-denmark-copyright-law-artificial-intelligence}, as viewed December 2025.}. Despite these efforts, most generative models continue to function as black boxes, offering limited visibility into how training data influence output. Even open-source models, while accessible, are too large to reproduce. Existing methods, such as influence functions or retraining-based attribution, require access to internal gradients or training pipelines, making them impractical for real-world black-box settings. Moreover, the reliance of some of them on latent representations limits the user's ability to understand how or why a particular image or style might have been replicated.

In this paper, we propose an interpretable framework for tracing training data influence through automatically generated KGs. Our method leverages multimodal LLMs to extract structured semantic information from images, represented as subject-predicate-object triples, and aligns these with a domain-specific ontology to construct KGs. These graphs serve as consistent and human-readable representations of image content, enabling semantic comparison between generated outputs and training samples.

By moving from parameter-dependent or embedding-based methods to comparisons over more structured and interpretable representations, our framework supports more transparent reasoning about what aspects of training data influence the generated output. For example, rather than stating that a generated image is similar to another on the embedding space, we can instead trace shared concepts such as objects, attributes, relationships, or stylistic motifs across KGs. This opens new possibilities for interpretable generation and real-world applications such as copyright auditing, dataset transparency, and artist-driven content curation.

The main contributions of this paper are as follows:

\begin{itemize}
    \item We propose a method to extract semantic triples from images using multimodal LLMs, aligning them with domain-specific ontologies to build interpretable KGs.
    \item We propose a graph-based influence analysis pipeline that compares generated and training images via graphs, enabling human-understandable explanations of similarity.
    \item We demonstrate the utility of our approach in settings involving copyright attribution, interpretability, and dataset transparency—highlighting how this type of reasoning can enhance traceability in generative models.
    \item We evaluate our method on locally trained models, and on large-scale models through a style-specific experiment. In the latter case, we focus on the recently popular Ghibli-style generations, to assess our method's ability to detect stylistic influence, even in black-box settings.\footnote{\url{https://apnews.com/article/studio-ghibli-chatgpt-images-hayao-miyazaki-openai-0f4cb487ec3042dd5b43ad47879b91f4}, as viewed December 2025.}
\end{itemize}

\paragraph{Paper Structure.} The paper is organised as follows: Section~\ref{sec:related-work} reviews related work on interpretability in generative models and KG construction using multimodal LLMs. Section~\ref{sec:method} presents our proposed method.  Section~\ref{sec:experiments} describes our experimental setups and results across local and large-scale generative models. Section~\ref{sec:conclusion} concludes with key findings, and outlines future directions.

\section{Related Work}
\label{sec:related-work}
\subsection{Interpretability in Generative Models}
As generative models advance, the need for interpretability has grown across technical, ethical, and policy domains. Broad methods such as t-SNE \cite{t-SNE}, LIME \cite{lime}, and attention visualisation \cite{attention} offer insights into model behaviour, but do not directly address training data attribution—an increasingly critical concern in applications involving authorship and copyright.
To understand how training data influences outputs, various attribution techniques have been proposed \cite{influencesurvey}. Early methods include influence functions \cite{koh}, retraining-based attribution \cite{black2021leaveoneoutunfairness}, and Shapley value approximations \cite{pmlrshapley}. Despite their usefulness and effectiveness in well-understood settings such as classification tasks, these approaches are often impractical for large-scale models due to their reliance on gradients, retraining access, or high computational cost. As a result, they are not applied in modern deep learning pipelines, especially in the context of generative models where access to internal parameters is unavailable. 

To overcome these limitations, model-agnostic methods have been proposed to compare generated outputs with training data in a shared embedding space, typically using features from pre-trained encoders, e.g. \cite{aivalis2025search}. While effective and scalable, these latent-space analyses rely on vector representations that are difficult to interpret. Recent work like LatentExplainer \cite{latentexplainer2024} attempts to bridge this gap by perturbing latent variables and using language models to describe their semantic effects. However, these methods still operate within an opaque space that limits transparency for end-users.

In contrast to latent or parametric methods, we use automatically generated KGs to trace data influence. By comparing generated and training images via shared triples, capturing objects, attributes, and stylistic elements, we enable scalable, interpretable, and human-aligned attribution.

\begin{figure*}[t]    
    \centering
    \includegraphics[width=1\linewidth]{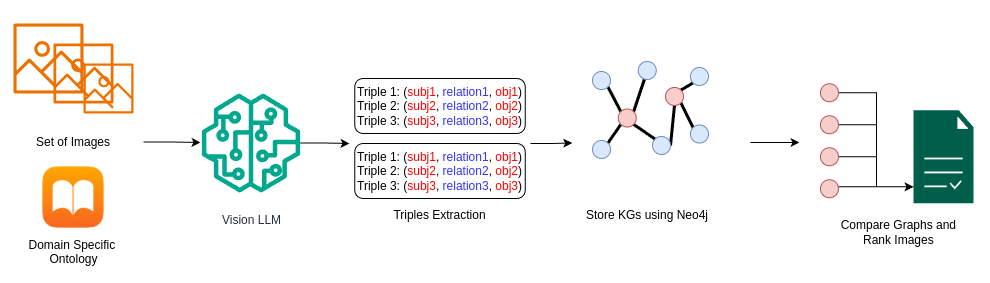}
    \caption{Pipeline overview. A multimodal LLM, prompted with a domain-specific ontology, extracts semantic triples from images. These are stored as per-image KGs and compared via node overlap to rank semantic similarity.}
    \label{fig:kg_pipeline}
\end{figure*}

\subsection{Large Language Models for Knowledge Graphs}
LLMs have become central in modern KG workflows, enabling structured reasoning over unstructured input data. Their primary contributions span three core areas: (1) generating  subject-predicate-object triples from natural text, (2) assisting in ontology creation, and (3) translating natural language inquiries into graph queries.

For triple extraction, prompt-based methods like AutoKG \cite{zhu2023llms} can convert text into ontology-aligned triples. Beyond extraction, LLMs have been leveraged for ontology learning and extension, using techniques such as prompt engineering, ontology reuse, and reasoning over specific concepts \cite{llms4life2024}. In question answering, LLMs can act as semantic parsers that generate SPARQL queries from user input. This is further improved by incorporating contextual KGs or using controlled natural languages to reduce ambiguity \cite{controlledkgqa2023,manufacturingkg2024}.
Surveys on generative KG construction \cite{kgcsurvey2023, multimodalkgsurvey2022} highlight the shift from traditional pipeline architectures toward unified, LLM-driven approaches. These systems handle entity recognition, relation extraction, and graph population, outperforming traditional pipelines.

These advances in KG construction provide a promising foundation for addressing the problem of data attribution in generative models. Instead of relying on opaque embeddings or inaccessible parameters, structured outputs from multimodal LLMs can be used to build interpretable KGs that capture key elements of both generated and training data and trace influences. Although most work so far focuses on text, these methods show strong potential for extension to images, despite the added complexity of visual content.

\section{Proposed Method}
\label{sec:method}
\subsection{Pipeline Overview}
In this paper, we propose a method to construct KGs from images using multimodal LLMs guided by domain-specific ontologies. The ontology, defining key semantic attributes and relations, is embedded in the prompt to ensure contextual relevance to the image and its domain.

Given an image, we prompt the LLM to generate structured triples aligned with the ontology. These triples capture the most salient objects, attributes, and relationships present in the image, and collectively form an interpretable KG. The structure of the resulting graph is entirely determined by the ontology itself. Each image produces its own interpretable graph, which we store in a graph database.
While each image has a distinct subgraph, many of the nodes—characteristics—are shared across images due to the ontology’s constrained vocabulary. As a result, all image-specific graphs are part of a larger, interconnected structure. This unified graph representation allows us to identify semantic overlap between images by comparing their subgraphs within the global knowledge space.
The full pipeline, including ontology-guided triple extraction, KG construction, and graph-based image ranking, is shown in Figure~\ref{fig:kg_pipeline}.

\subsection{Triples Extraction via Multimodal LLMs}
To describe the characteristics of each image, we use a multimodal LLM to extract structured semantic triples. These triples capture high-level concepts represented in the image.
The axes of interest are defined by a domain-specific ontology, designed by experts. The ontology serves as a semantic guide, ensuring that all images are described using a shared and consistent structure. To bridge the gap between the general-purpose LLM and the specialised ontology, we embed the ontology into the prompt during inference.

This approach enables the construction of interpretable per-image graphs and their integration into a unified global graph. Individual image graphs are treated as subgraphs, which are interconnected through shared attribute nodes, facilitating semantic comparison across the dataset.

\subsection{Graph Storage, Comparison, and Ranking}
Once semantic triples have been extracted, we store them in a graph database. This enables efficient querying, visualisation, and comparison across a dataset. The use of a domain-specific ontology ensures that all graphs follow a consistent schema, facilitating interpretable comparisons.
To identify similarities between images, we compare their graphs by measuring the number of shared nodes—typically attributes and concepts defined by the ontology. This node-level comparison captures semantic overlap between images and offers a more structured and interpretable approach compared to traditional feature-based or embedding-based similarity methods.

This graph-based retrieval process can be useful for uncovering insights about which training samples may have influenced a given generated image. By ranking training graphs based on their semantic similarity to the graph of a generated image, we aim to trace influence at a concept level rather than relying on opaque embeddings. The following section evaluates the approach for training data attribution in generative models.

We adopt a lightweight, node-level comparison to demonstrate the feasibility and interpretability of ontology-guided influence analysis in a scalable, model-agnostic setting.
In other words, the ontology is used primarily as a structured vocabulary to support the extraction of RDF triples from text.
The method does not perform reasoning or apply entailment rules based on the formal semantics of RDF(S) or OWL.
Expanding the framework to include semantic inference mechanisms is considered as future work.

\subsection{Scalability Considerations}
As will be shown in the next section, our framework was evaluated on large-scale models, though handling large datasets remains a common challenge for interpretability frameworks. Constructing ontology-aligned KGs for millions of images requires multimodal inference, but this process needs to be done once and be reused without repeating this stage. Each graph can be created within seconds and the procedure can be parallelised across multiple GPUs or distributed nodes. The resulting KGs serve as compact symbolic representations that occupy less storage and enable interpretable search, making the initial extraction cost a practical investment for transparent data exploration.


\begin{figure*}    
    \centering
    \includegraphics[width=1\linewidth]{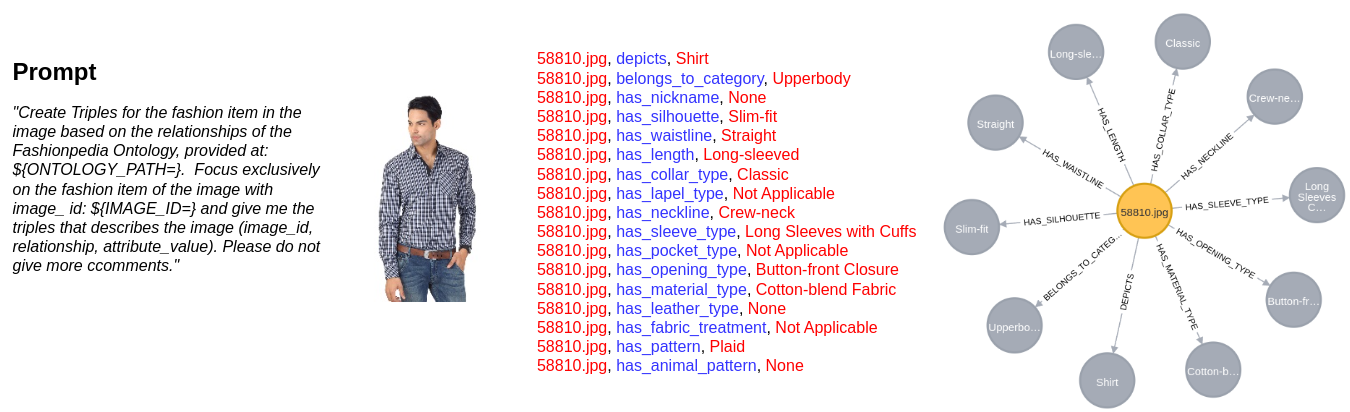}
    \caption{Example output of our system, which extracts semantic triples and constructs a KG from a fashion image. On the left, we show the prompt used to instruct the Multimodal LLM. On the right, the extracted knowledge is visualised as a graph, here, star-shaped due to the dataset showing a single garment per image.}
    \label{fig:prompt_example}
\end{figure*}

\section{Experiments}
\label{sec:experiments}
We evaluate our KG-based pipeline using two complementary setups for data attribution in generative models \footnote{Code available: 
\url{https://github.com/teoaivalis/AutoGraphX.git}}.
First, we conduct controlled data removal experiments on a locally trained model to assess whether our method can identify influential training samples. Second, we investigate stylistic influence in a black-box setting using Ghibli-style image generations. We construct a “Ghibli World” KG from original movies and compare it against stylised images. These setups allow us to examine both direct attribution and broader stylistic traceability in generative models.

\subsection{Experiment on a Locally trained Model}
To establish a fair and interpretable comparison, we reproduce the experimental setup proposed in a recent search-based influence framework  \cite{aivalis2025search}, which also operates without access to training parameters. This method operates in two stages: (1) it retrieves training samples whose textual descriptions are most similar to the user's prompt, and (2) ranks them based on cosine similarity between these samples and the generated output, via their embeddings. The most similar training images are then removed and the model is retrained—allowing for an empirical test of their influence.

We applied our graph-based method under the same conditions, using the same dataset, prompts, and generative model. Instead of relying on text and latent-space similarity, we retrieved influential samples based on the extracted KGs similarity. This setup allowed for a direct comparison between embedding-based and graph-based retrieval in terms of influence detection and interpretability.

\subsubsection{Experimental Setup}
\paragraph{Generative Model.} We used a small-scale generative model based on the open-source \texttt{dalle-pytorch} package\footnote{\url{https://github.com/lucidrains/DALLE-pytorch}, as viewed December 2025.}, trained on the Fashion Product Images \cite{fashionitems}. This is a dataset of 44,000 fashion images with accompanying captions. We retained the same configuration and hyperparameters as the original baseline. This setup allowed us to maintain full control over the training data, making it suitable for experimentation with data removal and retraining.

\paragraph{Multimodal LLM} After generating the images, we proceeded with the triple extraction process to semantically describe their content. We selected the open-source model LLaMA 3.2 Vision, accessed via the Ollama framework\footnote{\url{https://ollama.com/library/llama3.2-vision}, as viewed December 2025.}. This model was chosen as a practical solution capable of handling image inputs and producing structured outputs in natural language. Its ability to generate detailed, attribute-rich descriptions made it well-suited for transforming visual content into structured knowledge representations. 
Although our task focuses on extracting structured triples, it can be viewed as a form of image captioning. In this respect, recent evaluations of multimodal LLMs show comparable performance in visual understanding tasks \cite{verma2024mllm}. All models exhibit similar robustness trends. Since our contribution lies in how extracted knowledge is structured and compared we focus on one representative model.

\paragraph{Ontology.} To guide and constrain the triple extraction, we adopted the Fashionpedia ontology\footnote{\url{https://fashionpedia.github.io/home/}, as viewed December 2025.}, a comprehensive taxonomy developed for fashion-related visual understanding. It defines a structured set of object categories, parts, and attributes that are relevant to garments and accessories. By conditioning the multimodal LLM on this ontology during triple extraction, via the prompt, we ensured that the KGs followed a consistent, domain-specific structure. 

To assess the importance of the ontology, we also conducted a study in which triples were extracted without providing the ontology to the LLM. The results, computed after removing null values, are summarised in Table~\ref{tab:ontology-ablation}. On average, prompting with the ontology yields a significantly higher number of triples per image, covering more relationships. Without the ontology, the model tends to focus only on a few obvious visual cues, leading to lower coverage and limited ability to construct meaningful KGs. This illustrates that ontology guidance is critical for ensuring rich and consistent structures, which are essential for the graph-based comparison.

\begin{table}[h]
\centering
\caption{Ablation study on ontology-guided extraction: Statistical comparison of triples with and without ontology.}
\begin{tabular}{lcccc}
\toprule
\multicolumn{5}{c}{\textbf{Extracted Triples Statistics}} \\
\midrule
\textbf{Extraction Setting} & \textbf{Mean} & \textbf{Min} & \textbf{Max} & \textbf{Std. Dev.} \\
\midrule
Without Ontology & 3.44 & 1 & 20 & 1.38 \\
With Ontology    & 7.89 & 1 & 17 & 2.95 \\
\bottomrule
\end{tabular}
\label{tab:ontology-ablation}
\end{table}

The left part of Figure~\ref{fig:prompt_example} shows the prompt we use with an example of the extracted triples for a single training image, highlighting how different attributes are semantically structured. As shown, the output consists of multiple ontology-aligned triples covering various aspects of the garment, including category (\texttt{depicts}), shape (\texttt{has\_silhouette}), material (\texttt{has\_material\_type}), pattern (\texttt{has\_pattern}) and all the relationships defined by Fashionpedia ontology. Each triple follows the expected \texttt{(image\_id, relation, attribute)} format, providing an interpretable representation of the image. This level of structure enables downstream graph construction and comparison at the concept level.

\paragraph{Graph Construction and Ranking.} We stored all the triples in Neo4j\footnote{\url{https://neo4j.com/}, as viewed December 2025.}, a widely adopted database that supports efficient storage, querying, and visualisation of large-scale graph structures. For each image, we constructed a KG where the image ID is linked to a set of attributes (e.g., object type, material, pattern) defined by the ontology. In practice, the resulting graph are star-shaped due to the dataset showing a single garment per image. Each image contains a single garment without interacting entities. An example of this configuration is shown in right part of 
the Figure~\ref{fig:prompt_example}.
Attributes with missing or placeholder values (e.g., \texttt{None}, \texttt{N/A}) were excluded to reduce retrieval noise. All image-specific graphs were embedded into a unified global graph, where semantic attributes are shared across images and represented as common nodes.

To perform influence retrieval, we compared each generated image’s KG against those of all training images, computing node-level overlap. For each prompt, we retained the training images whose graphs shared the highest number of nodes with the graph of the generated one. These were used in the subsequent unlearning step. Beyond retrieval, Neo4j also enabled global exploration of the graph space—for instance, identifying the most frequent attributes or visualising the distribution of concepts across the dataset.


\paragraph{Unlearning Procedure.} We define unlearning as the process of retraining the generative model after removing a selected subset of training samples. This allows us to evaluate whether these samples had a possible impact on the generated output. We conducted 10 unlearning experiments using both the baseline and our method. In each experiment, we generated images for 15 prompts before and after unlearning. For consistency with the baseline setup of  \cite{aivalis2025search}, we removed a comparable proportion of training images corresponding to the most influential samples retrieved for each prompt.

After removing these samples, we retrained the model from scratch with identical hyperparameters and regenerated images for the same prompts. To measure the effect of unlearning, we compared each generated image before and after retraining using cosine similarity between their image embeddings, following the same evaluation protocol as the baseline.

Table~\ref{tab:unlearning} reports the mean, standard deviation, and range of cosine similarities across all experiments. As observed, the results from our graph-based method are generally comparable to those of the baseline. The similarity scores after unlearning remain close in magnitude, with small variations across runs. This consistency suggests that the graph-based pipeline is similarly effective at identifying influential training samples, even though it relies on structured, interpretable representations instead of latent embeddings.

\begin{table}
\centering
\caption{Cosine similarity across 10 experiments before/after unlearning. Our KG-based method matches baseline performance in influence detection.}
\scriptsize
\setlength{\tabcolsep}{3pt}
\renewcommand{\arraystretch}{1.5}
\begin{tabular}{c cccc @{\hspace{4pt}} cccc}
\toprule
& \multicolumn{4}{c}{\textbf{Experiments 1--5}} & \multicolumn{4}{c}{\textbf{Experiments 6--10}} \\
\cmidrule(lr){2-5} \cmidrule(lr){6-9}
& \textbf{Stage} & \textbf{Mean} & \textbf{Std} & \textbf{Range} & \textbf{Stage} & \textbf{Mean} & \textbf{Std} & \textbf{Range} \\
\midrule
& Before & 0.546 & 0.051 & 0.465–0.606 & Before & 0.531 & 0.056 & 0.453–0.614 \\
& After & 0.522 & 0.049 & 0.453–0.586 & After & 0.492 & 0.057 & 0.411–0.561 \\
& \begin{tabular}[c]{@{}c@{}}After\\(Graph)\end{tabular} & 0.521 & 0.057 & 0.440–0.599 & \begin{tabular}[c]{@{}c@{}}After\\(Graph)\end{tabular} & 0.499 & 0.058 & 0.410–0.572 \\
\cmidrule(lr){1-5} \cmidrule(lr){6-9}
& Before & 0.537 & 0.043 & 0.477–0.596 & Before & 0.553 & 0.062 & 0.459–0.631 \\
& After & 0.488 & 0.051 & 0.411–0.551 & After & 0.518 & 0.077 & 0.410–0.616 \\
& \begin{tabular}[c]{@{}c@{}}After\\(Graph)\end{tabular} & 0.494 & 0.055 & 0.411–0.573 & \begin{tabular}[c]{@{}c@{}}After\\(Graph)\end{tabular} & 0.520 & 0.076 & 0.419–0.636 \\
\cmidrule(lr){1-5} \cmidrule(lr){6-9}
& Before & 0.512 & 0.072 & 0.421–0.613 & Before & 0.523 & 0.070 & 0.419–0.613 \\
& After & 0.481 & 0.057 & 0.411–0.569 & After & 0.486 & 0.067 & 0.390–0.584 \\
& \begin{tabular}[c]{@{}c@{}}After\\(Graph)\end{tabular} & 0.481 & 0.061 & 0.405–0.581 & \begin{tabular}[c]{@{}c@{}}After\\(Graph)\end{tabular} & 0.486 & 0.055 & 0.412–0.561 \\
\cmidrule(lr){1-5} \cmidrule(lr){6-9}
& Before & 0.519 & 0.047 & 0.455–0.585 & Before & 0.544 & 0.068 & 0.451–0.633 \\
& After & 0.475 & 0.053 & 0.395–0.551 & After & 0.510 & 0.071 & 0.412–0.599 \\
& \begin{tabular}[c]{@{}c@{}}After\\(Graph)\end{tabular} & 0.487 & 0.054 & 0.427–0.577 & \begin{tabular}[c]{@{}c@{}}After\\(Graph)\end{tabular} & 0.503 & 0.078 & 0.403–0.609 \\
\cmidrule(lr){1-5} \cmidrule(lr){6-9}
& Before & 0.550 & 0.056 & 0.466–0.622 & Before & 0.526 & 0.060 & 0.439–0.605 \\
& After & 0.508 & 0.069 & 0.431–0.613 & After & 0.504 & 0.058 & 0.424–0.585 \\
& \begin{tabular}[c]{@{}c@{}}After\\(Graph)\end{tabular} & 0.508 & 0.080 & 0.385–0.612 & \begin{tabular}[c]{@{}c@{}}After\\(Graph)\end{tabular} & 0.499 & 0.058 & 0.427–0.594 \\
\bottomrule
\end{tabular}
\label{tab:unlearning}
\end{table}

\begin{figure*}    
    \centering
    \includegraphics[width=1\linewidth]{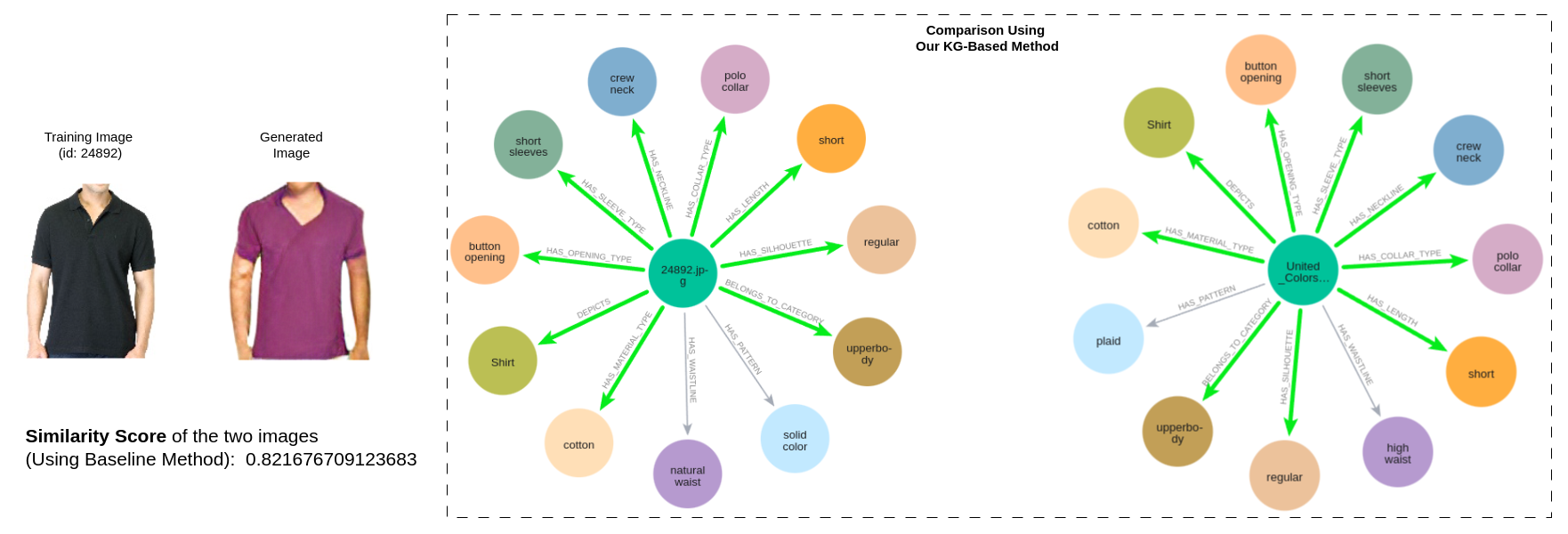}
    \caption{Comparison of similarity methods. Both identify the images as highly similar (baseline score: 0.82). The KG-based method also explains why by revealing shared attributes between the two images. Green edges indicate relationships whose attributes are identical across both graphs.}
    \label{fig:method_comparison}
\end{figure*}

These results align with our expectations: since embedding-based methods are optimised for image similarity, we anticipated that our graph-based approach might yield similar or slightly lower performance. The main advantage of our method lies in its interpretability. Unlike embeddings, which are difficult to reason about, our representations consist of human-readable attributes and relationships. This makes it easier to understand why certain training images are considered influential, as the influence can be traced to specific semantic elements rather than opaque vectors. As shown in Figure ~\ref{fig:method_comparison}, both the baseline and our method identify a training image as highly similar to the generated one. The key advantage of our approach is that it reveals the specific shared characteristics between the two images, whereas the baseline only produces a high similarity score without offering interpretability.

\subsubsection{Graph-Based Statistical Analysis of Retrieved Training Images}
While our unlearning experiments of the previous subsection measured the influence of retrieved training samples by comparing the similarity of generated images before and after retraining, these results alone do not fully explain why certain images may have had such an effect. To complement that evaluation, we now turn to a graph-level statistical analysis of the retrieved samples.
Specifically, we analysed the structure and semantic characteristics of the top-ranked training images selected per prompt by our graph-based method. This analysis aims to reveal whether the retrieved samples form coherent, semantically dense groups.
Table~\ref{tab:graph-stats} summarises core graph statistics across 10 unlearning experiments (each comprising 15 prompts), providing insight into the consistency and clustering of attributes within each retrieved group. The results help validate the effectiveness of our retrieval strategy and offer an interpretable view into what makes these samples influential in the generation.

\begin{table*}
\centering
\caption{Graph statistics of retrieved images across experiments E1–E10. High relation diversity and large attribute clusters indicate that our method captures meaningful, coherent visual concepts aligned with human understanding.}
\label{tab:graph-stats}
\begin{tabular}{lccccc}
\toprule
\textbf{Statistic} & \textbf{E1} & \textbf{E2} & \textbf{E3} & \textbf{E4} & \textbf{E5} \\
\midrule
Distinct Relationships   & 14.80±1.64 & 15.36±0.89 & 14.93±1.18 & 15.21±1.32 & 15.00±1.55 \\
Avg. Unique Values/Rel.  & 4.76±1.09  & 4.62±0.89  & 3.93±0.84  & 4.87±0.58  & 4.09±1.11 \\
Avg. Shared Attributes   & 5.15±2.11  & 6.03±1.65  & 6.91±1.63  & 5.78±1.23  & 6.00±1.79 \\
Max Shared Attributes    & 10.07±2.11 & 13.36±8.4 & 13.40±8.06 & 13.50±8.42 & 10.13±2.22 \\
Largest Cluster ($\geq$5)     & 35.80±6.22 & 38.79±3.05 & 40.00±0.00 & 39.21±1.82 & 37.00±9.25 \\
Largest Cluster ($\geq$7)     & 23.33±13.86 & 34.29±8.79 & 36.80±5.05 & 31.71±9.87 & 30.60±12.30 \\
\midrule
\textbf{Statistic} & \textbf{E6} & \textbf{E7} & \textbf{E8} & \textbf{E9} & \textbf{E10} \\
\midrule
Distinct Relationships   & 15.33±2.15 & 15.8±1.42 & 15.47±0.96 & 15.6±1.14 & 15.2±1.47 \\
Avg. Unique Values/Rel.  & 4.00±1.23 & 4.93±1.63 & 4.48±1.03 & 5.08±1.11 & 4.74±1.72 \\
Avg. Shared Attributes   & 5.03±1.69 & 6.82±2.13 & 6.12±1.96 & 5.36±1.53 & 7.14±1.99 \\
Max Shared Attributes    & 9.73±2.52 & 10.87±1.59 & 10.73±2.24 & 10.93±1.39 & 11.2±1.64 \\
Largest Cluster ($\geq$5)     & 33.73±9.51 & 39.6±0.80 & 37.4±6.57 & 37.47±5.69 & 40±0 \\
Largest Cluster ($\geq$7)     & 23.6±14.61 & 33.87±10.20 & 30.67±12.82 & 30.2±10.4 & 33.67±9.18 \\
\bottomrule
\end{tabular}
\end{table*}

\paragraph{Distinct Relationships}
Across all experiments, the retrieved graphs consistently contain a rich set of relationships (mean $\approx$ 15 per group). The low standard deviation across experiments (typically $<$ 1.5) suggests that the ontology-guided extraction process is highly stable. This indicates that the multimodal LLM, when guided by the ontology, can extract diverse and semantically rich information from each image. A high number of distinct relationships ensures that the generated KGs are not overly repetitive or trivial. Importantly, this supports the idea that this representation capture meaningful structure—going beyond simple visual similarity.

\paragraph{Unique Values per Relationship}
The number of unique values per relationship averaged between 3.9 and 5.1 across all experiments. This moderate diversity is well-suited to a controlled domain like fashion, where variation exists but bounded by ontology categories (e.g., limited silhouettes or patterns). The values show that retrieved samples are similar enough to form coherent groups, yet diverse enough to avoid redundancy. This ensures that structured influence arises from shared high-level concepts and not duplication.

\paragraph{Shared Attributes: Depth of Semantic Similarity}
The average number of shared attributes across retrieved graphs ranges from 5 to 7.1, while the maximum values extend up to 13 attributes in some groups. This suggests that retrieved images are not only semantically rich but also semantically aligned with one another. Prompts that include more structured cues typically lead to higher average shared attributes, while more abstract or stylistic prompts yield greater variation. This shows the system’s ability to adapt to prompt specificity and surface semantically aligned examples.

\paragraph{Clustering Patterns: Evidence of Semantic Grouping}
In nearly all experiments, the retrieved samples formed large clusters of semantically overlapping graphs. Clusters with $\geq$5 shared attributes consistently had mean sizes between 34 and 40, with standard deviations below 6. This indicates that most top-ranked training samples are conceptually close, not just in isolated attributes but in overall graph structure. Clusters with $\geq$7 shared attributes, though naturally smaller, still exhibited average sizes in the range of 23–37. This pattern highlights the retrieval method’s ability to identify not just influential individuals but coherent subgroups of training data.

\paragraph{Cross-Experiment Consistency and Conclusion}
The low variance across most metrics demonstrates that the graph extraction and retrieval pipeline performs consistently across different prompts and experimental seeds. This consistency is critical for interpretability: if the system retrieved semantically inconsistent or noisy groups, it would fail to offer meaningful insights about influence. Instead, we observe strong clustering, stable relationship diversity, and reliable attribute sharing. These findings confirm that our ontology-grounded pipeline surfaces training images that are not only visually or textually similar but also semantically aligned with the generated outputs. In summary, the statistical analysis supports our core claim: automatically constructed KGs provide a transparent, interpretable, and structured foundation for assessing training data influence in generative models—advancing the goals of explainability, copyright auditing, and dataset transparency.

\subsection{Stylised Domain Case Study}
\label{sec:stylised_domain}
\subsubsection{Motivation and Challenges}
The recent surge in stylised image generation, particularly following the release of advanced multimodal capabilities within the ChatGPT platform, highlighted a growing trend towards Ghibli-style outputs. This ``Ghibli Effect'' increased user activity, with reports indicating over one million new users joining ChatGPT in a single hour\footnote{\url{https://www.reuters.com/technology/artificial-intelligence/ghibli-effect-chatgpt-usage-hits-record-after-rollout-viral-feature-2025-04-01/}, as viewed December 2025.}. Users adopted the generation of personal or fictional scenes in the Ghibli aesthetic, raising challenges regarding attribution, originality, and influence.

When prompting a generative model to create an image in a specific artistic style---such as ``Ghibli-style''---the process does not involve the direct reproduction of specific training images. Instead, the model learns the general visual characteristics of the style during training. These characteristics, stored in the model's parameters, are extracted from a collection of images that represent the style, but are not directly copied.

As a consequence, attribution becomes complex. There is no one-to-one correspondence between training samples and the generated outputs. Influence emerges at a higher level of abstraction---through shared motifs, textures, themes, and compositional structures. Understanding stylistic influence therefore requires a different experimentation compared to direct object-level attribution, focusing on structured representations and semantic comparison.

\subsubsection{Experimental Setup}
\begin{figure}
    \centering
    \includegraphics[width=0.5\textwidth]{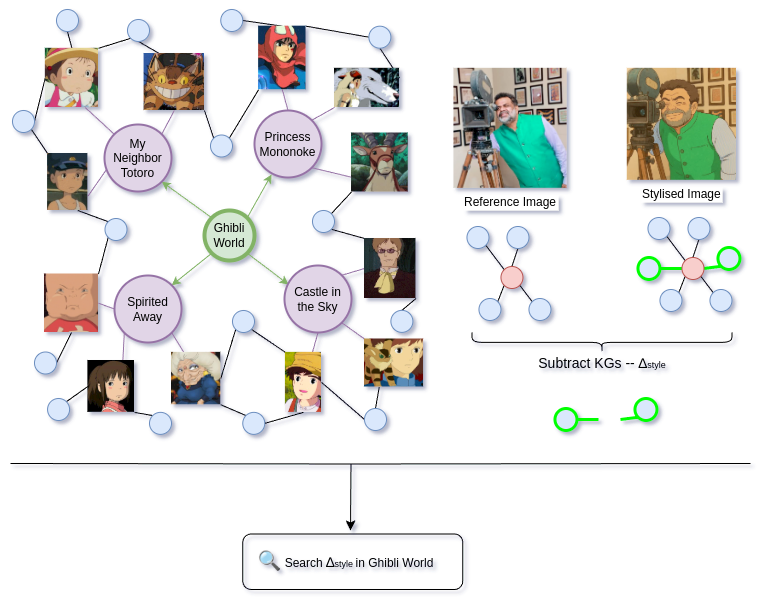}
    \caption{Experimental pipeline overview. The left illustrates the ``Ghibli World'' from movies. On the right, we can find the $\Delta_{\text{style}}$. Blue nodes represent semantic characteristics.}
    \label{fig:ghibli_pipeline}
\end{figure}

To investigate stylistic influence, we build upon the Automatic KG generation pipeline of the previous sections. Our experiment is structured as follows:

\begin{enumerate}
    \item \textbf{Ghibli Movie Frames:} We begin with a collection of frames extracted from popular Studio Ghibli films \cite{ghibli_image_pairs}. Each frame is described using a domain-specific ontology by extracting structured semantic triples. These are used to construct a ``Ghibli World'' KG that aims to capture the key elements of the studio’s visual language. Although our dataset contains only a subset of Ghibli’s released films, it includes ten of the most iconic and widely referenced titles \footnote{\url{https://editorial.rottentomatoes.com/guide/all-studio-ghibli-movies-ranked-by-tomatometer/}, as viewed December 2025.}, supporting the assumption that the selected frames sufficiently reflect the studio’s representative style.
    
    \item \textbf{Reference and Styled Image Pairs:} In parallel, we use a dataset containing pairs of real-world reference images and their corresponding Ghibli-style versions \cite{ghibli_movies}. We apply the same ontology-guided extraction process to describe both the original and the stylised images as structured KGs.

    \item \textbf{Style-Induced Difference:} For each image pair, we subtract the KG of the reference image from that of the stylised one to isolate the elements introduced from generation, defined as $\Delta_{\text{style}} = \text{KG}_{\text{stylised}} - \text{KG}_{\text{reference}}$. The resulting set of new nodes, attributes, and relationships represents the candidate stylistic elements to be evaluated against the Ghibli World.
    
    \item \textbf{Matching Against the Ghibli World:} Finally, we compare the style-induced elements from each image with the Ghibli World. This step enables us to assess whether the characteristics introduced during generation are consistent with the visual semantics of the original films. When matches occur, we further trace them to specific source movies within the Ghibli World, offering evidence that the model has absorbed stylistic patterns from particular films. More broadly, this comparison serves as a validation of our systems ability to capture the core visual characteristics of the Ghibli aesthetic.
\end{enumerate}
Figure~\ref{fig:ghibli_pipeline} summarises the process, showing the data sources (image pairs and movie frames), the KG extraction, and the final comparison used to assess stylistic influence.

\subsubsection{Knowledge Graph Extraction}
To extract semantic information from images, we employ a multimodal-LLM (LLaMA 3.2 Vision). The extraction process is guided by a domain-specific ontology to ensure consistency and semantic alignment across triples.
As a starting point, we adopted the ``Lightweight Ontology for Describing Images'' (LIO)~\footnote{\url{https://lov.linkeddata.es/dataset/lov/vocabs/lio}, as viewed December 2025.}, which provides a minimal structure for representing visual content. LIO defines core relationships such as \textit{depicts}, \textit{shows}, \textit{conveys}, \textit{usesPictorially}, and \textit{hasSetting}, among others.

To capture the stylistic characteristics relevant to our study, particularly texture, colour, mood, lighting, and compositional patterns, we extended the initial ontology with object properties and attributes. This allowed us to describe both depicted entities and stylistic elements introduced during generation.
Table~\ref{tab:ontology_relationships} summarises the allowed properties used during the extraction process.

\setlength{\tabcolsep}{3pt} 

\begin{table}
\centering
\caption{Ontology relationships used for KG extraction.}
\begin{tabular}{@{}ll@{}}
\toprule
\textbf{Category} & \textbf{Allowed Relationships} \\
\midrule
General Semantics & depicts, shows, conveys, looksLike \\
\addlinespace
Stylistic Features & usesPictorially, hasArtisticElement \\
\addlinespace
Scene Composition & hasInForeground, hasInBackground, \\
                   & hasDepictedBackground, \\ 
                   & hasPictorialBackground \\
\addlinespace
Contextual Setting & hasSetting \\
\addlinespace
Visual Qualities & hasColorPalette, hasLineQuality, \\
                 & hasTexture, hasLightingEffect \\
\addlinespace
Character,                   & hasCharacterDesign, \\
Environment Style            & hasEnvironmentStyle \\
\addlinespace
Mood and Atmosphere & hasMoodAtmosphere \\
\addlinespace
Composition Details & hasCompositionType \\
\addlinespace
Attribute Properties & style, technique, materials, hasTag, \\ 
                        & styleDetails \\
\bottomrule
\end{tabular}
\label{tab:ontology_relationships}
\end{table}

\subsubsection{Results and Analysis}
In total, we analysed 1,519 style-induced triples by subtracting the KGs of reference images from their corresponding stylised versions. These triples represented new semantic elements introduced during generation. We then matched these elements against the ``Ghibli World'' KG, which was constructed from the original movie frames.
Of the 1,519 triples, only 40 triples (2.63\%) introduced attributes not found in the frames, suggesting a high degree of alignment between the stylistic output of the generative model and the visual characteristics of Studio Ghibli films. 
Manual inspection revealed that most of these unmatched triples corresponded to paraphrased concepts, reflecting linguistic variability rather than conceptual divergence.

To further trace stylistic influence, we recorded the source movie of each matched triple. Aggregating these matches showed that most films contributed similar amounts of stylistic elements, with \textit{Princess Mononoke}, \textit{Kiki's Delivery Service}, and \textit{The Tale of Princess Kaguya} having slightly higher counts. This indicates that the Ghibli World provides a broad coverage of the stylistic patterns reflected in the generated images. The full distribution is shown in Table~\ref{tab:ghibli_influence_table}.

\subsubsection{Discussion}
This case study illustrates that stylistic influence can be analysed using structured representations. By extracting and comparing KGs from reference and stylised images, and aligning them with a curated ``Ghibli World'' graph, we were able to identify which stylistic elements were introduced during generation and trace them back to specific source films.
Importantly, the distribution of matched triples across Ghibli movies was relatively even, indicating that the pipeline reflects a broad stylistic understanding rather than copying from a small number of examples. This reinforces the value of our graph-based approach: it offers a transparent and interpretable way to assess stylistic influence in black-box generative systems, supporting more informed analysis in areas such as attribution, copyright, and ethical AI auditing.

\begin{table}
\centering
\caption{Ghibli movies contributing matched stylistic triples.}
\begin{tabular}{lcc}
\toprule
\textbf{Movie} & \textbf{Triples} & \textbf{Percentage (\%)} \\
\midrule
Princess Mononoke & 11,447 & 10.99 \\
Kiki's Delivery Service & 11,345 & 10.89 \\
The Tale of Princess Kaguya & 11,080 & 10.64 \\
Spirited Away & 10,815 & 10.39 \\
Castle in the Sky & 10,261 & 9.85 \\
My Neighbor Totoro & 10,166 & 9.76 \\
Grave of the Fireflies & 10,034 & 9.64 \\
Whisper of the Heart & 9,864 & 9.47 \\
Howl's Moving Castle & 9,629 & 9.25 \\
Nausicaa of the Valley of the Wind & 9,492 & 9.12 \\
\bottomrule
\end{tabular}
\label{tab:ghibli_influence_table}
\end{table}

\section{Conclusions and Future Work}
\label{sec:conclusion}
We introduced a framework for interpreting generative models by constructing ontology-aligned KGs from images using multimodal LLMs. Our method enables semantic comparison between generated and training samples without relying on internal model or opaque embeddings. By using structured representations, we offer an interpretable approach for tracing potential influences in image generation.
Through experiments on locally and large-scale black-box models, we demonstrated that KGs can support both direct data attribution and broader stylistic analysis. In our unlearning experiments, graph-based retrieval achieved comparable influence detection to latent-space approaches, while offering improved interpretability. In our stylistic case study, we showed that the model’s outputs could be linked to specific Ghibli films, highlighting the ability of structured reasoning to capture abstract stylistic influence.

Our results indicate that ontology-driven KGs provide a practical and interpretable approach to data attribution in generative AI. Furthermore, through automatically generating KGs, our pipeline is potentially usable across domains and disciplines where transparency and interpretability are required.

Looking forward, several directions remain open. First, while our experiments focused on fashion and stylised visual domains, future work could explore more diverse and general-purpose datasets. In particular, applying our framework to large-scale datasets would allow us to evaluate its scalability in even more realistic settings and investigate distributed or parallelised extraction strategies. This would allow for further validation of triple extraction guided by a domain ontology, and its ability to capture meaningful semantics.
Second, our similarity metric currently relies on node-level overlap between graphs. More expressive comparison methods e.g., edge-weighted similarity, could enhance retrieval precision and reveal deeper relationships between training data and generative outputs. We also plan to extend the framework with semantic inference capabilities, supporting richer and more semantically informed retrieval. For artistic analysis, incorporating additional contextual information—such as metadata about artists, stylistic schools, or social influences—could enrich the graph representations and enable more detailed comparisons of artistic works.

Third, improving user interpretability remains a key priority. We aim to develop interactive tools for exploring graph-based comparisons, enabling content creators, legal experts, and model developers to trace influence in a visual and accessible way.
Finally, we plan to collaborate with domain experts to refine and extend the ontologies used in graph construction. Their input will help ensure that the extracted knowledge remains both technically robust and meaningful within specialised application contexts.


\printbibliography

\end{document}